\documentclass[conference]{IEEEtran}
\IEEEoverridecommandlockouts
\usepackage[utf8]{inputenc} 
\usepackage{cite}
\usepackage{amsmath,amssymb,amsfonts}
\usepackage{algorithm}
\usepackage{algpseudocode}
\usepackage{graphicx}
\usepackage{textcomp}
\usepackage{xcolor}
\usepackage{caption}
\usepackage{float}
\usepackage{subcaption}
\usepackage{psfrag}
\usepackage{url}
\usepackage{wrapfig}
\usepackage[colorlinks=false]{hyperref}

\def\BibTeX{{\rm B\kern-.05em{\sc i\kern-.025em b}\kern-.08em
    T\kern-.1667em\lower.7ex\hbox{E}\kern-.125emX}}

\usepackage{cite}
\usepackage{amsmath,amssymb,amsfonts}
\usepackage{algorithm}
\usepackage{algpseudocode}
\usepackage{graphicx}
\usepackage{textcomp}
\usepackage{xcolor}
\usepackage{caption}
\usepackage{float}
\usepackage{subcaption}
\usepackage{url}
\usepackage{wrapfig}
\usepackage{hyperref}
\usepackage[utf8]{inputenc}

\title{\LARGE \bf
Custom Non-Linear Model Predictive Control for Obstacle Avoidance in Indoor and Outdoor Environments*
}

\author{Lara Laban$^{1,2}$, Mariusz Wzorek$^{3}$, Piotr Rudol$^{3}$ and Tommy Persson$^{3}$
\thanks{*This work was supported by the project UAS@LU Autonomous Flight and WASP. The LU author is member of ELLIIT.}
\thanks{$^{1}$The author is with the Department of Automatic Control, Lund University (LU), Lund, Sweden 
{\tt\footnotesize lara.laban@control.lth.se}}%
\thanks{$^{2}$The author is with Lund University School of Aviation (LUSA).}%
\thanks{$^{3}$The authors are with the Department of Computer Science, Linköping University (LiU) and WARA Public Safety (WARA-PS), Linköping, 
Sweden {\tt\footnotesize\{mariusz.wzorek,piotr.rudol,tommy.persson\}@liu.se}}%
}

\begin{document}

\maketitle
\thispagestyle{empty}
\pagestyle{empty}

\begin{abstract}

Navigating complex environments requires Unmanned Aerial Vehicles (UAVs) and autonomous systems to perform  trajectory tracking and obstacle avoidance in real-time. While many control strategies have effectively utilized linear approximations, addressing the non-linear dynamics of UAV, especially in obstacle-dense environments, remains a key challenge that requires further research. This paper introduces a Non-linear Model Predictive Control (NMPC) framework for the DJI Matrice 100, addressing these challenges by using a dynamic model and B-spline interpolation for smooth reference trajectories, ensuring minimal deviation while respecting safety constraints. The framework supports various trajectory types and employs a penalty-based cost function for control accuracy in tight maneuvers. The framework utilizes CasADi for efficient real-time optimization, enabling the UAV to maintain robust operation even under tight computational constraints. Simulation and real-world indoor and outdoor experiments demonstrated the NMPC ability to adapt to disturbances, resulting in smooth, collision-free navigation.

\end{abstract}
\vspace{-0.025cm}
\section{INTRODUCTION}

\begin{figure}[!t]
    \vspace{0.2cm} 
    \centering
    \captionsetup{font=footnotesize}
    \includegraphics[width=0.477\textwidth]{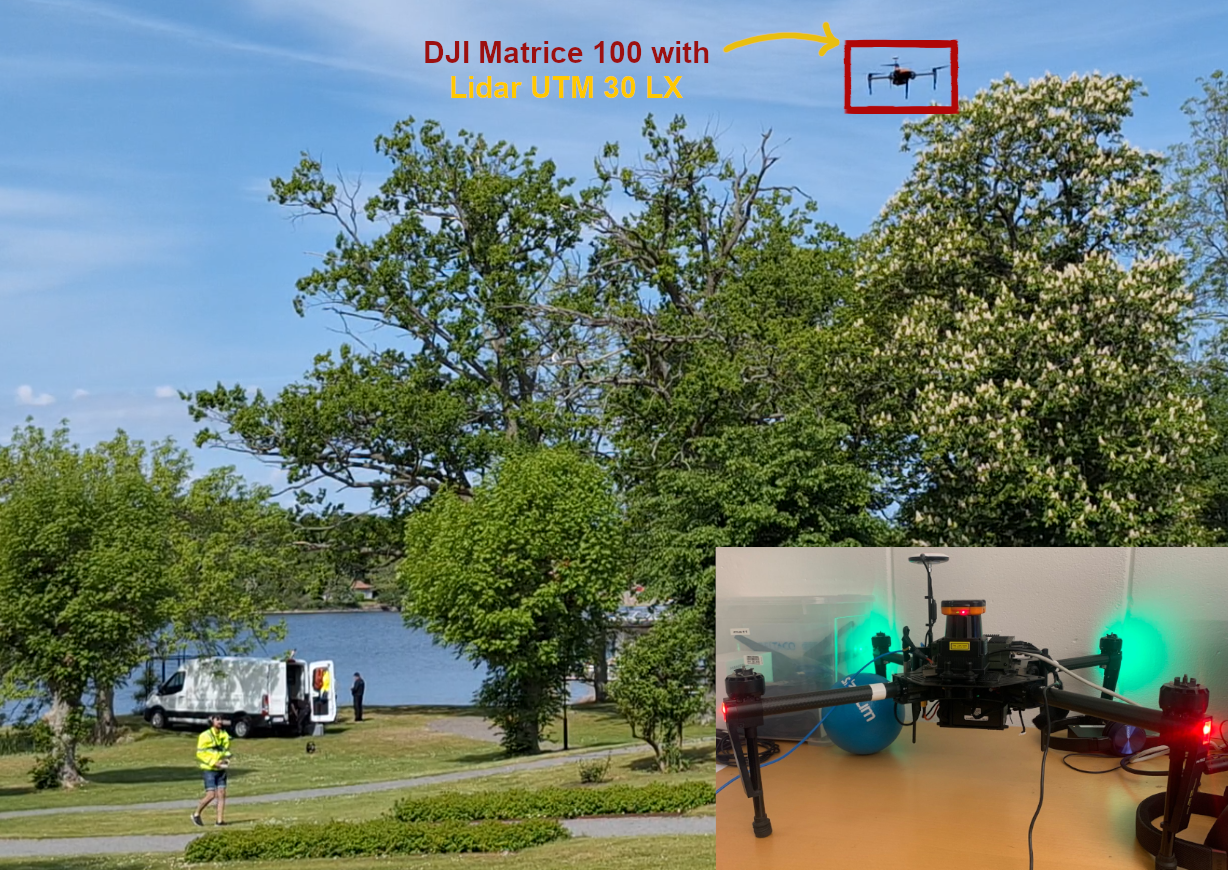}
    \caption{Experimental flight of DJI Matrice 100 demonstrating non-linear model predictive control for obstacle avoidance at Gränsö 2024.}
    \label{fig:piotr_dron}
    \vspace{-0.6cm}
\end{figure}

In recent years, Unmanned Aerial Vehicle (UAV) usage has surged, fueled by advances in control systems and computational power. A major challenge in autonomous UAV operations is implementing strategies such as Non-linear Model Predictive Control (NMPC) \cite{rawlings2017model}, which solves an optimization problem to find control inputs in complex environments (Fig.~\ref{fig:piotr_dron}). Various NMPC frameworks have been proposed, including one that combines linear MPC with nonlinear feedback to achieve agile maneuvers in dynamic, multi-vehicle scenarios without pre-planned paths \cite{baca2018model}. Another framework, using Bayesian Policy Optimization efficiently handles dynamic obstacles, combining stochastic optimization and MPC to enable real-time collision avoidance in unpredictable environments \cite{andersson2016model}. Additionally, deep learning-based approaches have improved real-time control, achieving faster and safer navigation through risk-aware active learning, with up to 50 times faster computation \cite{andersson2017deep}. Another approach integrates sensor constraints and actuator limitations to improve control in complex environments, showcasing NMPC adaptability to multi-rotor systems \cite{jacquet2020perception}. Further, NMPC frameworks have been designed for specific applications, such as stabilizing UAVs near barriers while reducing power consumption by up to $12.5\%$ \cite{kocer2019aerial}, optimizing path-following in wind-disturbed conditions, and improving urban target tracking accuracy with a gimbaled camera for small fixed-wing aircrafts \cite{chen2021horizontal, tyagi2021nmpc}. For example, NMPC has been used in tracking and releasing a cable-suspended load, showing effectiveness in real-time \cite{panetsos2024nmpc}. To ensure safe operations in cluttered environments, another NMPC framework has integrated a deep neural network for collision avoidance using depth images \cite{jacquet2024nmpc}. This approach allows autonomous systems to anticipate and avoid obstacles effectively, increasing safety and reliability. In multi-vehicle formations, NMPC frameworks maintain spatial geometries even with communication delays, ensuring reliability \cite{erskine2021model}. 

\begin{figure*}[t]
    \centering
    \vspace{-0.2cm}
    \captionsetup{font=footnotesize}
    \includegraphics[width=1.0\textwidth]{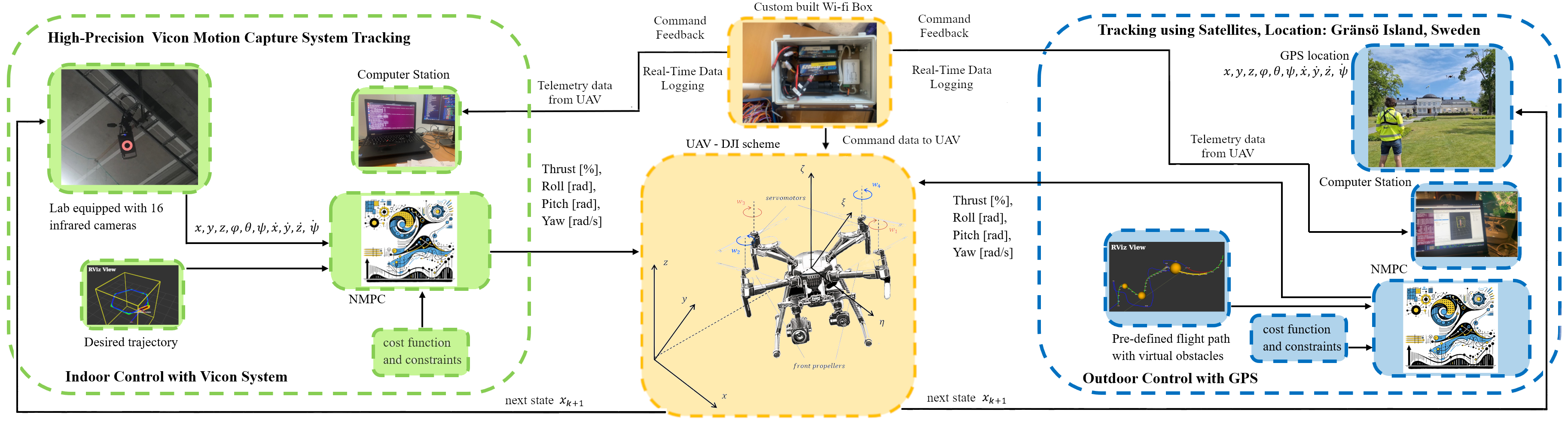}
    \caption{Control System Architecture: On the left, a Vicon motion capture system with 16 cameras captures 6-DOF data, which is processed by a computer station running NMPC to generate control commands for the UAV. On the right, the outdoor system uses GPS tracking at Gränsö, with NMPC calculating control inputs based on GPS data. The figure illustrates the body-fixed frame \((x, y, z)\) and the rotational directions of the four propellers \((w_1, w_2, w_3, w_4)\), along with the axes of rotation \((\xi, \eta, \zeta)\).}
    \vspace{-0.4cm}
    \label{fig:control_arch}
\end{figure*}
However, while these frameworks provide valuable solutions, many are tailored for smaller UAVs or specific applications, leaving gaps in general obstacle avoidance and scalability for larger UAVs. In other related work, NMPC frameworks for smaller Micro Aerial Vehicles (MAVs) have been shown to handle full system dynamics well under disturbances \cite{kamel2017linear}, but these approaches can face limitations in larger UAVs because of computational demands and more complex dynamics. Similarly, obstacle prediction methods, such as those using the PANOC solver \cite{lindqvist2020nonlinear}, are effective in real-time environments with moving obstacles, but their scalability is limited, especially on platforms with constrained onboard processing power. Additionally, while NMPC combined with visual servoing has been effective in specialized tasks like photovoltaic array inspection \cite{velasco2024visual}, such frameworks are narrowly focused and do not necessarily address the broader challenges of general obstacle avoidance in dynamical, real-world conditions. This paper extends these previous efforts by focusing on general obstacle avoidance for large UAVs while focusing on computational efficiency. 

\setcounter{footnote}{3}
\subsection{Problem Statement and Paper Contributions}

Here, we address the problem of obstacle avoidance in UAV flight control by means of model predictive control. NMPC maintains system nonlinearity, enabling the UAV to handle large deviations effectively. Building on \cite{kamel2017linear}, we validated an NMPC framework tailored to the DJI Matrice 100, with offloading used to manage computational demands. We applied Python CasADi \cite{casadi} for optimization and integrated it with Robot Operating System (ROS) \cite{ros2024}, providing an open-source code repository for UAV path planning in simulations and real-world tests \cite{my_github}.
\section{METHODS}

This section defines the control framework, including the quadrotor model \cite{mellinger2011minimum} and customized NMPC algorithms \cite{kamel2017linear} for obstacle avoidance and B-spline trajectory tracking.

\subsection{Quadrotor Dynamics}

The quadrotor is modeled as a rigid body with six degrees of freedom: three translational and three rotational (Fig.~\ref{fig:control_arch}). The state vector \(\mathbf{x}\) includes the UAV position, orientation angles, and velocities, while the control input vector \(\mathbf{u}\) consists of thrust \(T\) generated by the rotors and the reference angular rates for roll \(\phi_{r}\), pitch \(\theta_{r}\), and yaw rate \(\dot{\psi}_{r}\) \cite{bouabdallah2004design}.

\subsubsection{Translational Dynamics}

The UAV motion in \(x\), \(y\), and \(z\) directions is influenced by thrust, gravity, and aerodynamic drag, and is described by the following equation\cite{mellinger2011minimum},

\begin{equation}
\label{jednacina_jedan}
\begin{aligned}
\mathbf{F} = \frac{1}{m} \mathbf{R} \cdot  \mathbf{T} + \mathbf{g} - \mathbf{D}
\end{aligned}
\end{equation}
\noindent
where \(\mathbf{R}\) is the rotation matrix from body frame to inertial frame, \(\mathbf{T} = [0, 0, T]^T\) is the thrust vector in the body frame, \(\mathbf{g} = [0, 0, -g]^T\) and \(\mathbf{D} = -[b_x \dot{x}, b_y \dot{y}, b_z \dot{z}]^T\), are the gravity vector and drag forces in the inertial frame, respectively. In other works, such as \cite{bouabdallah2004design}, translational dynamics are often linearized around a hover point to simplify computation, though this may reduce accuracy during fast maneuvers. Keeping Eq. (\ref{jednacina_jedan}) nonlinear ensures the UAV full translational dynamic response is captured, enabling NMPC to handle large deviations and complex trajectories \cite{pereira2019}, \cite{mellinger2012trajectory}.

\subsubsection{Rotational Dynamics}

The UAV orientation is parameterized by the Euler angles \(\phi, \theta, \psi\) \cite{bouabdallah2007full} with angular acceleration determined by inputs, moments, and inertia, 
\begin{equation}
\begin{aligned}
\mathbf{\tau} = \mathbf{I} \cdot \dot{\mathbf{\omega}} + \mathbf{\omega} \times (\mathbf{I} \cdot \mathbf{\omega})
\end{aligned}
\end{equation}
\noindent
where \(\mathbf{\tau}\) denotes the total applied torques, \(\mathbf{I}\) is the inertia matrix, \(\mathbf{\omega}\) is the angular velocity, and \(\dot{\mathbf{\omega}}\) is the angular acceleration. In the cascaded system, the DJI Matrice 100's internal controllers operate in a closed-loop configuration, stabilizing the UAV pitch, roll, and yaw.
At the higher level, the NMPC computes the optimal control inputs for following a desired trajectory, handling the more complex, slower dynamics of translational movement \cite{bouabdallah2004design, lindqvist2020nonlinear, aerospace2020}. Eq. (\ref{trojka}) provides a linear approximation of the closed-loop rotational dynamics, enabling real-time NMPC computation.
The model structure was chosen based on the approach from \cite{kamel2017linear}, but the step response experiments were conducted on the DJI Matrice 100 in this work.\footnote{Direct tuning of the system was constrained by the proprietary nature of the DJI Matrice 100's internal controllers, limiting fully modifying the control parameters.} 
NMPC dynamically optimizes inputs over a prediction horizon for robust performance \cite{rawlings2017model}. The closed-loop dynamics of roll, pitch, and yaw are approximated by first-order systems with time constants and gains determined by the internal controllers (adapted based on the model in \cite{kamel2017linear}),

\begin{subequations}
\begin{align}
\label{trojka}
    \dot{\phi} &= \frac{1}{\tau_{\phi}} k_{\phi} (\phi_{r} - \phi) \\
    \dot{\theta} &= \frac{1}{\tau_{\theta}} k_{\theta} (\theta_{r} - \theta) \\
    \ddot{\psi} &=  \frac{1}{\tau_{\psi}} k_{\psi} (\dot{\psi}_{r} - \dot{\psi})
\end{align}
\end{subequations}
\noindent
where \(\tau_{\phi}, \tau_{\theta}\) are the time constants for roll and pitch, \(k_{\phi}, k_{\theta}\) are the proportional gains for roll and pitch, \(\tau_{\psi}\) is the time constant, and  \(k_{\psi}\) is the gain for the yaw rate.

\subsubsection{Custom NMPC Development for UAV}
\label{subsec:controller_formulation}

The quadrotor dynamics are described by differential equations describing the system motion, with the state vector \(\mathbf{x}\) reduced to 10 states as a result of the closed-loop dynamics of the UAV,
\vspace{-0.15cm}
\begin{subequations}
\begin{align}
\mathbf{x} &= [x, y, z, \phi, \theta, \psi, \dot{x}, \dot{y}, \dot{z}, \dot{\psi}]^T \label{eq:state_vector} \\
\mathbf{u} &= [T, \phi_{r}, \theta_{r}, \dot{\psi}_{r}]^T \label{eq:control_input}
\end{align}
\end{subequations}
\noindent
where, \(x, y, z\) are positions, \(\phi, \theta, \psi\) are roll, pitch, and yaw angles, \(\dot{x}, \dot{y}, \dot{z}\) are the linear velocities, and \(\dot{\psi}\) is the yaw rate. The control input vector is given in Eq. (\ref{eq:control_input}) where \(T\) is the total thrust, \(\phi_{r}\), \(\theta_{r}\) are roll and pitch reference angles and \(\dot{\psi}_{r}\) is the yaw rate reference.

\subsection{Optimization Problem}
\label{subsec:optimization_problem}

The optimization problem in the NMPC is solved at each sampling time, and only the first control input \(\mathbf{u}_0\) is applied to the system. The horizon shifts forward, the state is updated, and the optimization repeats.\footnote{We used multiple shooting for the discretization to solve the boundary value problem with continuity constraints\cite{bock1984multiple}.} The cost function \(J\), designed to minimize deviations from the reference trajectory and control inputs while penalizing excessive control rates, is defined as,

\begin{equation}
\begin{aligned}
J = & \sum_{k=0}^{N-1} \bigg[ \|\mathbf{x}_{k} - \mathbf{x}_{r,k}\|_Q^2 + \|\mathbf{u}_{k} - \mathbf{u}_{r,k}\|_R^2 \\
    & + \|\Delta \mathbf{u}_k\|_{R_{\Delta}}^2 \bigg] + \|\mathbf{x}_N - \mathbf{x}_{r,N}\|_{Q_f}^2
\end{aligned}
\end{equation}
\noindent
where \(\mathbf{x}_k\) and \(\mathbf{u}_k\) are the state and control input, indexed by the time step \(k\), with \(\mathbf{x}_{r,k}\) and \(\mathbf{u}_{r,k}\) as their respective references. Here, \(Q\), \(R\), and \(Q_f\) are the weighting matrices for state, control input, and final state deviation, respectively, while \(R_{\Delta}\) weights the rate of change of control inputs and \(\Delta \mathbf{u}_k = \mathbf{u}_k - \mathbf{u}_{k-1}\) encourages smooth control actions. The following equation and inequality represent both the state update dynamics and the control input constraints,
\begin{equation}
\begin{aligned}
\vspace{-0.2cm}
\mathbf{x}_{k+1} = \mathbf{x}_k + \mathbf{f}(\mathbf{x}_k, \mathbf{u}_k) \cdot dt, \quad \mathbf{u}_{min} \leq \mathbf{u}_k \leq \mathbf{u}_{max}
\end{aligned}
\end{equation}
\noindent
where \(\mathbf{f}(\mathbf{x}_k, \mathbf{u}_k)\) is the system dynamics function, and \(dt\)  is the discretization time step.

\subsection{Obstacle Incorporation in NMPC}
\label{subsec:obstacle_incorporation}

Obstacle penalty is added to the cost function for computational efficiency with a safety distance incorporated to reduce collision risks \cite{warren1989global}, acknowledging that avoidance may not always be achieved, further discussed in Sec.~\ref{realworld_section}. Obstacles are represented as spheres with a safety volume. The penalty term for each obstacle \(i\) is expressed as:
\begin{equation}
\begin{aligned}
d(\mathbf{x}_k, \mathbf{c}_i) \geq r_i + d_s, \quad \forall k \in \{0, ..., N\}
\end{aligned}
\end{equation}
\noindent
where \(d(\mathbf{x}_k, \mathbf{c}_i)\) is the Euclidean distance between the UAV position at time step \(k\) and the center of obstacle \(i\), \(r_i\) represents the radius of the obstacle and \(d_i\) is the safety distance around the obstacle. To achieve obstacle avoidance within the NMPC framework, a penalty term is added to the objective function, which increases as the UAV nears an obstacle. This penalty is formulated using a repulsive potential field approach \cite{khatib1986real}, introducing a strong repulsive force when the UAV gets too close. The repulsive potential \(U_i\) for obstacle \(i\) is defined as,
\begin{equation}
\begin{aligned}
\label{eq:eta_rolf}
U_i(\mathbf{x}_k) = \begin{cases}
\frac{1}{2} \eta \left( \frac{1}{d(\mathbf{x}_k, \mathbf{c}_i)} - \frac{1}{r_i + d_s} \right)^2 & \text{if } d(\mathbf{x}_k, \mathbf{c}_i) < r_i + d_s \\
0 & \text{otherwise}
\end{cases}
\end{aligned}
\end{equation}
\noindent
where \(\eta\) is a scaling factor determining the strength of the repulsive force. The total repulsive potential \(U(\mathbf{x}_k)\) from all obstacles is added to the NMPC objective function,
\begin{equation}
\begin{aligned}
J_{\text{total}}(\mathbf{x}, \mathbf{u}) = J + \sum_{i} U_i(\mathbf{x}_k)
\end{aligned}
\end{equation}

\subsection{Path Representation with B-splines}

B-splines were chosen for path representation because they generated smooth, adaptable paths based on provided waypoints, forming a trajectory that does not necessarily pass through each waypoint but still follows the overall intended path \cite{xu2023vision}. The B-spline curve is defined by control points and constructed using basis functions \( N_{i,p}(t) \) over a knot vector \( \mathbf{T} \) \cite{deboor2001bspline}. For degree \( p = 0 \), \( N_{i,0}(t) \) is 1 if \( t_i \leq t < t_{i+1} \) and 0 otherwise. For higher degrees, \( N_{i,p}(t) \) is a weighted combination of the previous degree's basis functions. The curve representing the reference path is a smooth, flexible path formed by a linear combination of these basis functions weighted by the control points.

\section{EXPERIMENTAL SETUP}

\subsubsection{Technical Specifications}


\begin{figure}[htbp]
\vspace{-0.3cm}
    \centering
    \captionsetup{font=footnotesize}
    \begin{tabular}{@{\hspace{-0.9mm}}c@{\hspace{-0.9mm}}c}
        \begin{minipage}[b]{0.248\textwidth}
            \centering
             \raisebox{2.0mm}{\includegraphics[width=\textwidth]{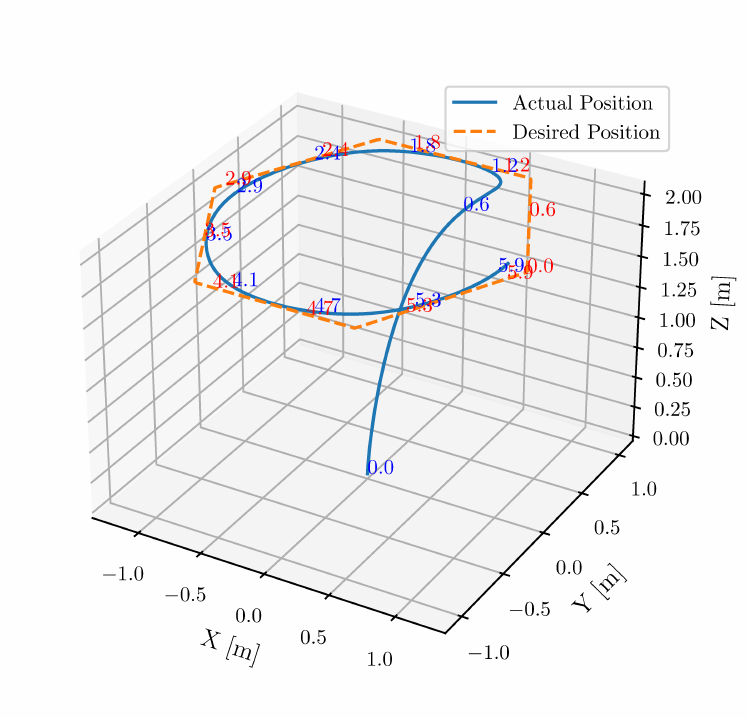}}
            \label{fig:simulator_rviz_multi_obstacle}
        \end{minipage} &
        \begin{minipage}[b]{0.248\textwidth}
            \centering
           {\includegraphics[width=\textwidth]{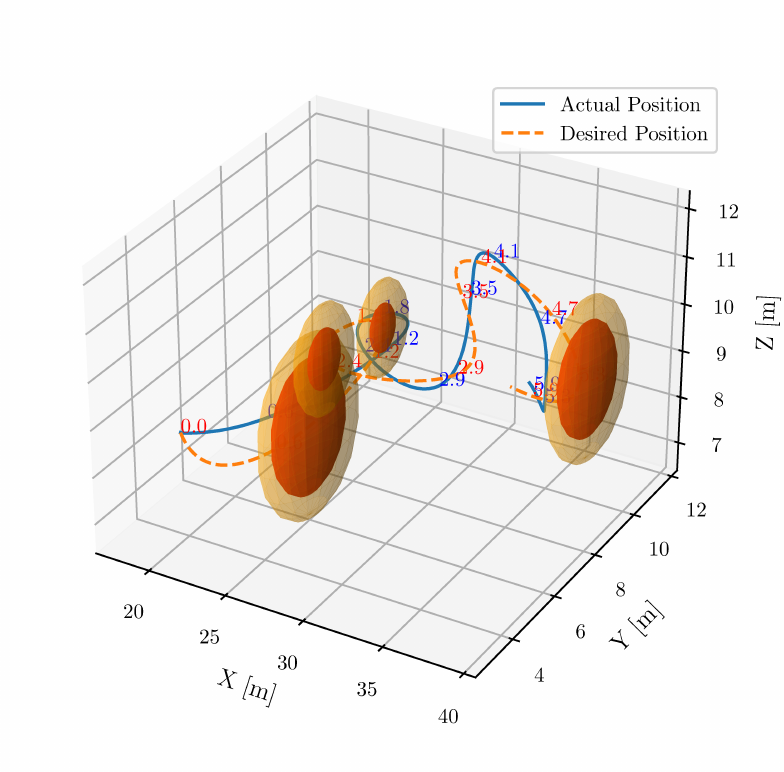}}
            \label{fig:bspline_path_obstacles_hexagon}
        \end{minipage}
    \end{tabular}
    \vspace{-0.3cm}
    \caption{Desired and Actual Path Following in 3D with NMPC in the Hexagonal and Multiple Obstacle Scenario Simulations: The left image depicts a hexagonal path formed using a B-spline, the right image the multiple obstacle path, where the blue line represents the actual flight path, the yellow dashed line indicates the desired trajectory, and the red spheres depict the obstacles, surrounded by a yellow safety distance.}
    \label{fig:combined_simulations}
    \vspace{-0.4cm}
\end{figure}

The NMPC algorithm was evaluated on a DJI Matrice 100 UAV equipped with an Intel NUC computer\footnote{\scriptsize{\href{https://www.intel.com/content/www/us/en/content-details/334661/7th-generation-intel-processor-families-for-u-y-platforms-datasheet-volume-1-of-2.html?wapkw=Intel\%20NUC7\%20i7-7567U}{\texttt{https://www.intel.com/content/www/us/en/content-details/
content-details/334661/7th-generation-intel-processor [Accessed: September 15, 2024].}}}} (Intel® Core™ i7-7567U CPU at 3.50GHz, 16GB RAM, 500GB SSD storage) ensuring high-performance computation essential for real-time control tasks. The onboard NUC runs Ubuntu 20.04.6 LTS, integrating ROS, which serves as the basis for managing all UAV operations, including communication, control, and data processing.\footnote{The NMPC problem was implemented in Python using the CasADi library \cite{andersson2019casadi}, which supports symbolic computation and numerical optimization. The IPOPT solver \cite{casadi} was used to solve the nonlinear programming (NLP) problem, with mumps/ma27 employed as the linear solver.} 

\begin{table}[htbp]
\caption{Summary of Performance Metrics from Closed-Loop Simulations}
\begin{center}
\footnotesize 
\begin{tabular}{|p{3.8cm}|p{1.6cm}|p{1.9cm}|}
\hline
\textbf{Metric}                          & \textbf{Hexagonal Path}            & \textbf{Multi-Obstacle Avoidance}       \\ \hline
\textbf{Average Deviation [m]}      & 0.21                               & 1.01                              \\ \hline
\textbf{Maximum Deviation [m]}      & 2.33                               &  2.85                               \\ \hline
\textbf{Avg. No. of Solver Iterations}     & 2.11                                 & 4.06                                 \\ \hline
\textbf{Avg. Convergence Time [s]} & 0.02                      & 0.03                             \\ \hline
\textbf{Input Control Range (Thrust)}    & [37.14, 37.15]                     & [34.85, 39.87]                    \\ \hline
\textbf{Total Time for Scenario [s]}     & 27.50                               & 28.57                            \\ \hline
\end{tabular}
\label{tab:simulation_metrics}
\end{center}
\vspace{-0.4cm}
\end{table}

\subsubsection{Indoor Experimental Configuration}

Indoor experiments were conducted in a controlled environment using the Vicon Motion Capture System \cite{vicon} for real-time precise UAV tracking. The controller was tested under various scenarios to assess its adaptability and robustness, including low-speed precise movements and higher-speed operations. The NMPC controller was tuned with a weight matrix \( \mathbf{R} = \text{diag}([0.2, 0.5, 0.5, 0.1]) \), which provided the best performance during the tuning process.\footnote{This configuration prioritized less oscillations in roll and pitch, addressing previous issues with thrust and yaw oscillations by reducing the emphasis on thrust and yaw while improvingcontrol along the more critical roll and pitch axes.}

\begin{figure}[htbp]
    \centering
    \captionsetup{font=footnotesize}
    \begin{tabular}{@{\hspace{-0.6mm}}c@{\hspace{-0.5mm}}c}
        \begin{minipage}[b]{0.248\textwidth}
            \centering
        \hspace{0.5cm}\includegraphics[width=0.85\textwidth]{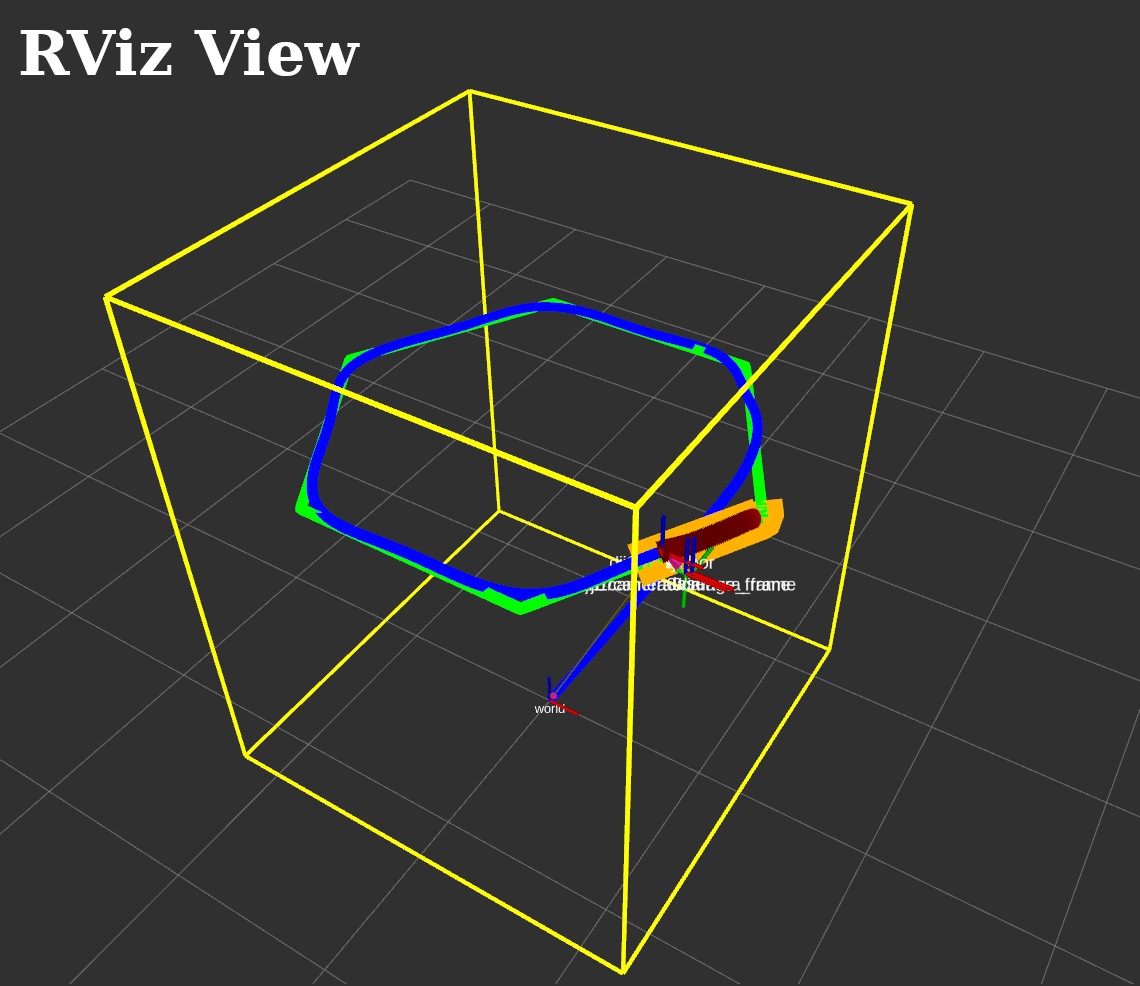}
        \end{minipage} &
        \begin{minipage}[b]{0.248\textwidth}
            \centering
            \includegraphics[width=\textwidth]{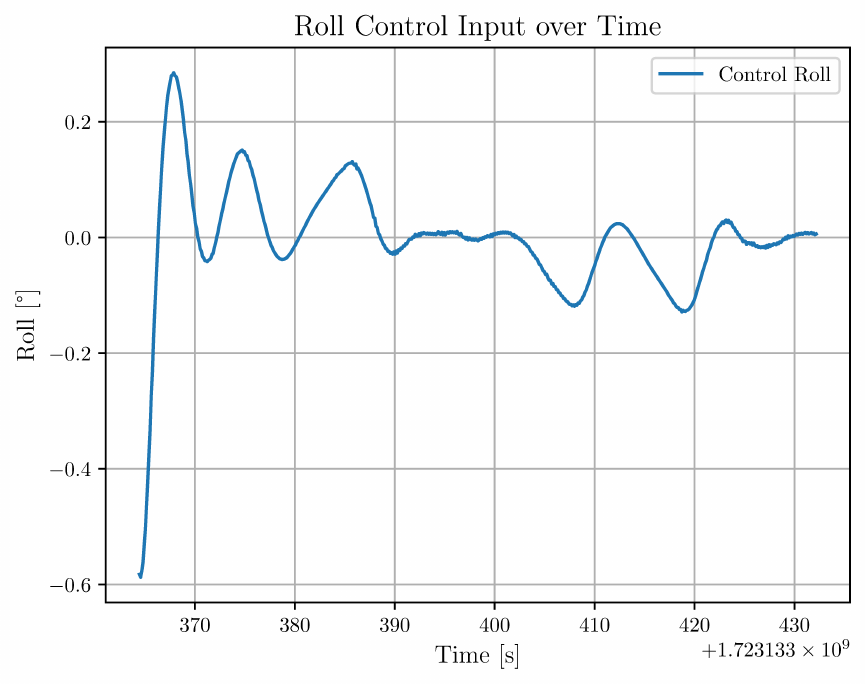}
        \end{minipage} \\
        \begin{minipage}[b]{0.248\textwidth}
            \centering
            \includegraphics[width=\textwidth]{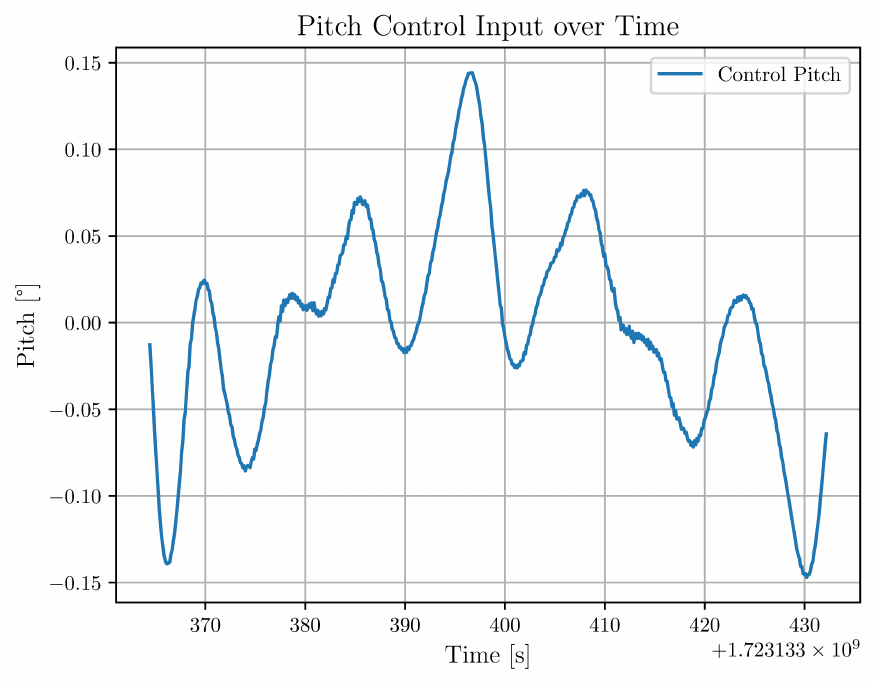}
        \end{minipage} &
        \begin{minipage}[b]{0.248\textwidth}
            \centering
            \includegraphics[width=\textwidth]{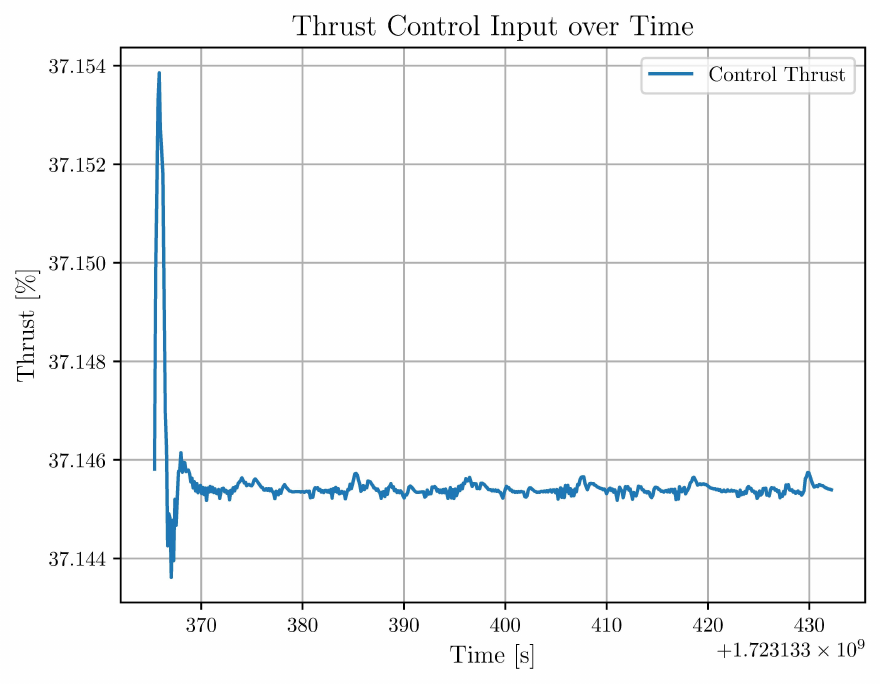}
        \end{minipage} \\
        \begin{minipage}[b]{0.248\textwidth}
            \centering
            \includegraphics[width=\textwidth]{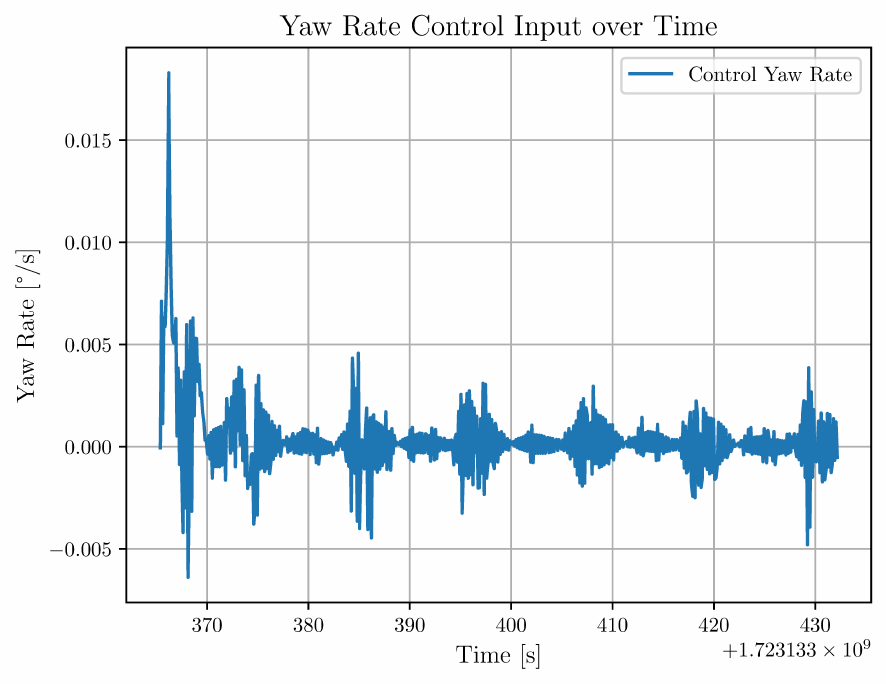}
        \end{minipage}  &
        \begin{minipage}[b]{0.248\textwidth}
            \centering
            \includegraphics[width=\textwidth]{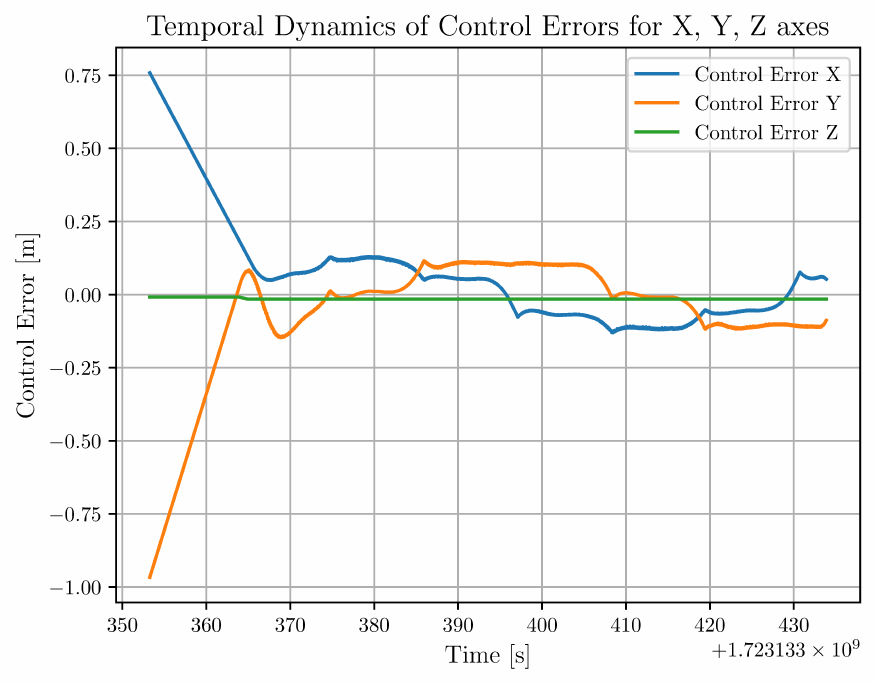}
        \end{minipage} 
    \end{tabular}
    \caption{Comparative Analysis of UAV Path Following with NMPC in the Hexagonal Path Scenario: In the upper left corner the yellow square outlines the Vicon system's operational limits, indicating the area prepared for indoor experiments, the green line is the desired trajectory, the blue line is the actual path taken by the UAV, the red line shows the prediction horizon, and the yellow line indicates the path being sent to the controller at the current time step. The control inputs for thrust, roll, pitch, and yaw rate are shown in the lower sections, alongside control error plots for the X, Y, and Z axes (This applies to Figs.~\ref{fig:combined_figure_multi_obstacle_4_simulator_results},~\ref{fig:nmpc_path_tracking_obstacle}, and~\ref{fig:combined_figure_multi_obstacle_3} as well).}
    \vspace{-0.29cm}
    \label{fig:hexagon}
\end{figure}

\subsubsection{Outdoor Experimental Configuration}

Outdoor experiments challenged the NMPC controller with wind disturbances and unstructured environments. The UAV navigated an obstacle course, requiring real-time adjustments to avoid collisions. After the indoor tests, 3 outdoor tests were conducted, tuning \( \mathbf{R} = \text{diag}([0.3, 0.6, 0.6, 0.2]) \) to reduce oscillations and in order to compensate for the wind. Tuning was guided by error analysis and real-time RViz plots to ensure smoother responses in dynamic outdoor conditions.

\begin{table}[htbp]
\caption{Summary of Performance Metrics from Simulated HIL Flight Experiments}
\begin{center}
\footnotesize 
\begin{tabular}{|p{4.1cm}|p{1.2cm}|p{1.9cm}|}
\hline
\textbf{Metric}                          & \textbf{Hexagonal Path} & \textbf{Multi-Obstacle Avoidance} \\ \hline
\textbf{Navigation Time [s]}             & 80.60                    & 44.50                                \\ \hline
\textbf{Navigation Time (obstacles) [s]} & N/A               & 56.78                                 \\ \hline
\textbf{Time Increase (obstacles) [\%]} & N/A                & 27.60                              \\ \hline
\textbf{Average Deviation [m]}           &  0.19                     & 1.52                                  \\ \hline
\textbf{Average Deviation (obstacles) [m]} & N/A             &  1.87                                 \\ \hline
\textbf{Average Solver Iterations}               & 2.00                  & 5.60                                  \\ \hline
\textbf{Avg. Solver Iterations (obstacles)} & N/A               & 7.52                                   \\ \hline
\end{tabular}
\label{tab:simulated_gps_flight_metrics}
\vspace{-0.4cm}
\end{center}
\end{table}

\subsection{Open Source Code and Video Documentation}

The open-source code, along with a user manual, can be found in the GitHub repository \cite{my_github}. In addition, videos from Gränsö Testings 2024 are provided \cite{video_granso}.

\section{EXPERIMENTS AND RESULTS}

\subsection{Simulation Results}

\subsubsection{Closed-Loop Simulation and Analysis}
\label{simulation_section}

Closed-loop simulations were conducted to evaluate the NMPC controller performance. Figure~\ref{fig:combined_simulations} illustrates the actual and desired trajectories under different scenarios, including obstacle avoidance and following predefined paths. The same NMPC model was used for both the simulation and the simulated HIL flight tests. The control loop update times ranged from approximately $\mathrm{0.033}$ $\mathrm{s}$ to $\mathrm{0.011}$ $\mathrm{s}$, depending on the computational load during real-time processing. This variability allowed the NMPC controller to make timely adjustments, providing smooth and accurate trajectory tracking, as summarized in Table~\ref{tab:simulation_metrics}.

\subsubsection{Simulated Flight Experiment with Hardware-in-the-Loop (HIL)}
\label{HIL_section}
\begin{figure}[H]
    \centering
    \captionsetup{font=footnotesize}
    \begin{tabular}{@{\hspace{-0.6mm}}c@{\hspace{-0.5mm}}c}
        \begin{minipage}[b]{0.248\textwidth}
            \centering
        \includegraphics[width=\textwidth]{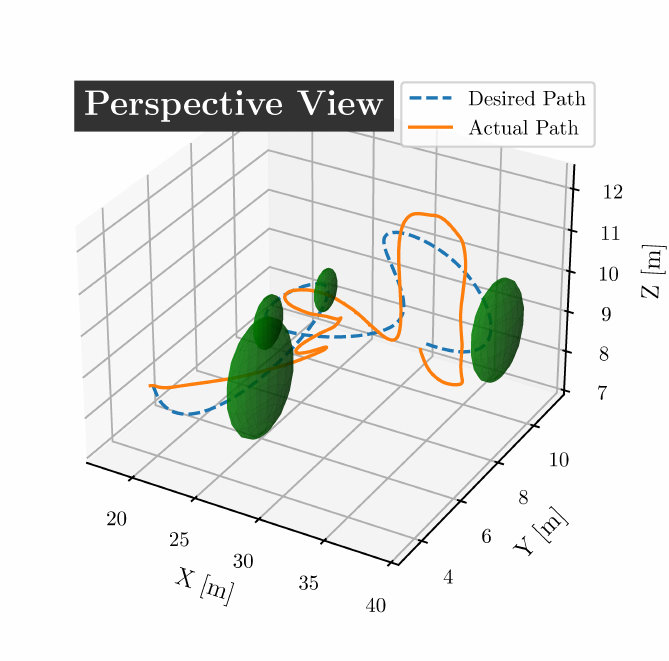}
        \end{minipage} &
        \begin{minipage}[b]{0.248\textwidth}
            \centering
        \includegraphics[width=\textwidth]{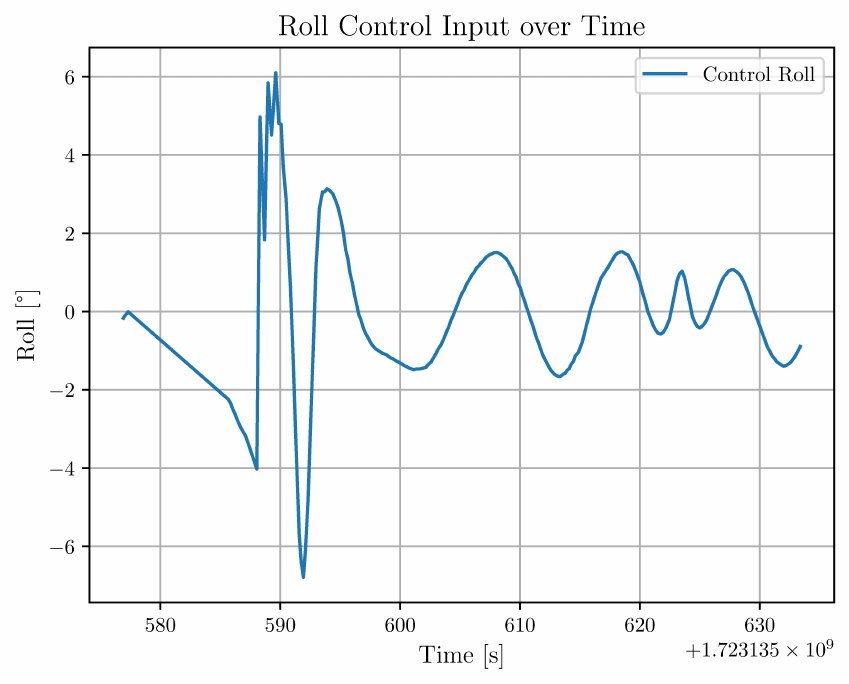}
        \end{minipage} \\
        \begin{minipage}[b]{0.248\textwidth}
            \centering
            \includegraphics[width=\textwidth]{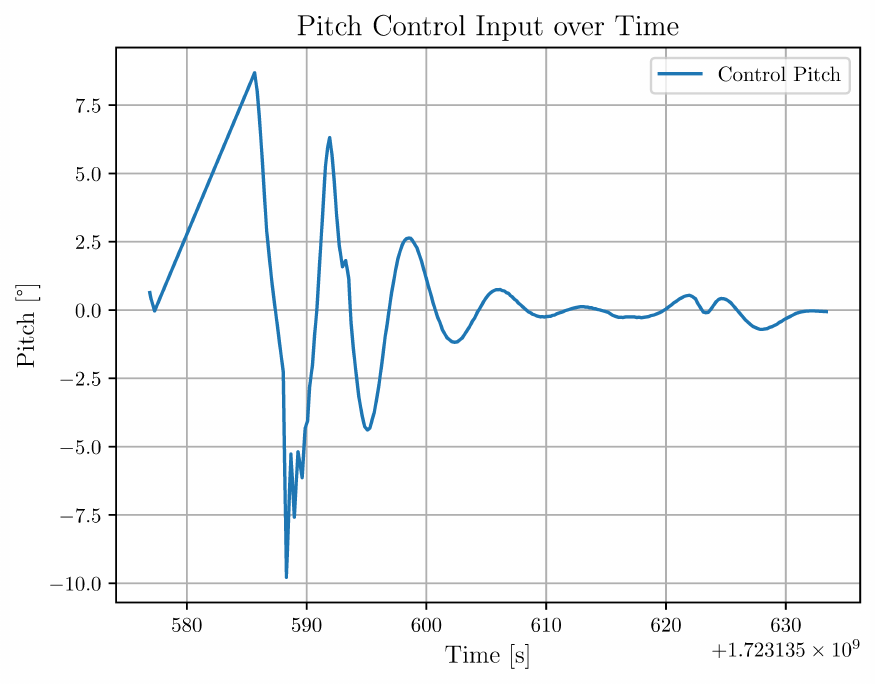}
        \end{minipage} &
        \begin{minipage}[b]{0.248\textwidth}
            \centering
        \includegraphics[width=\textwidth]{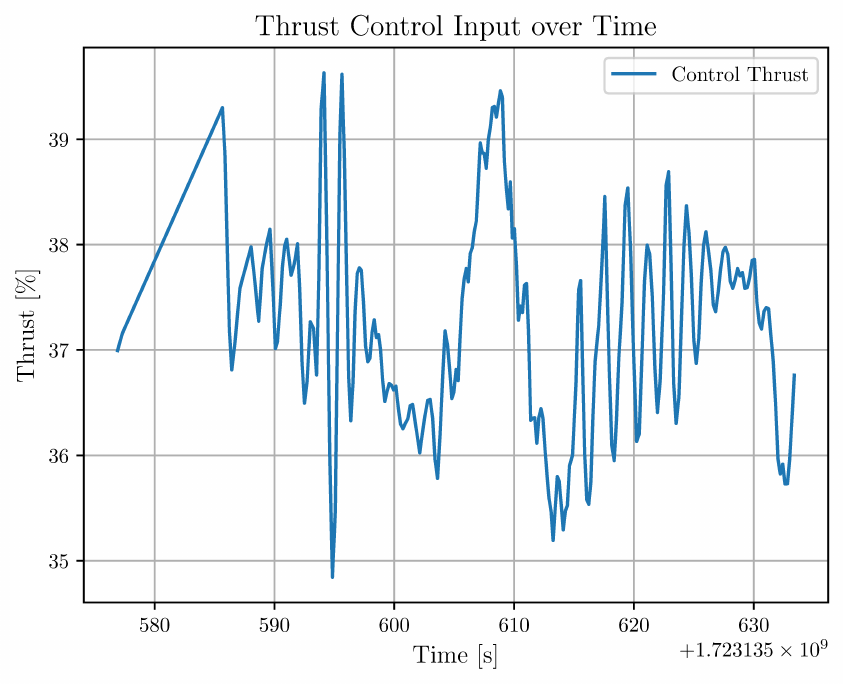}
        \end{minipage} \\
        \begin{minipage}[b]{0.248\textwidth}
            \centering
        \includegraphics[width=\textwidth]{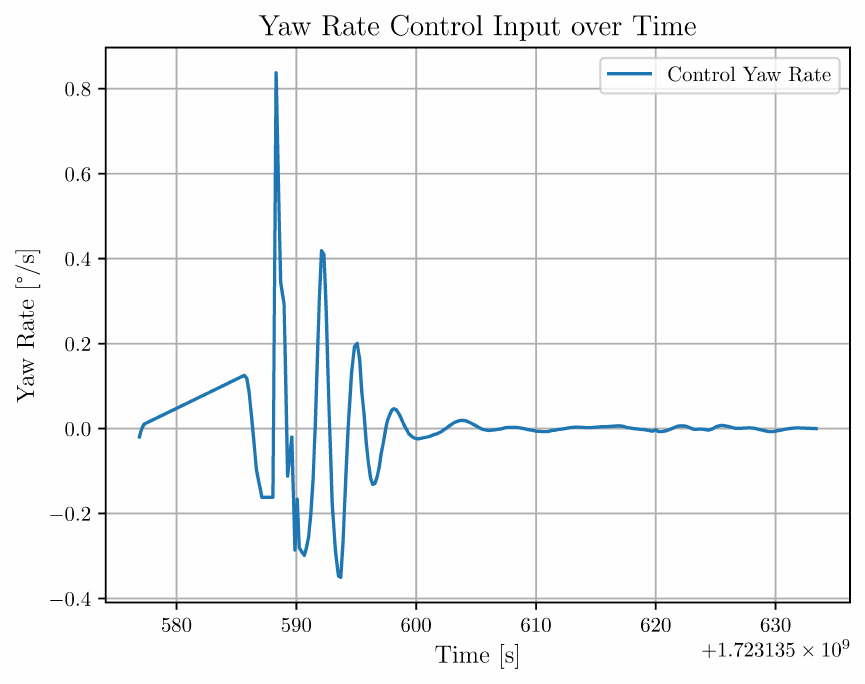}
        \end{minipage} &
        \begin{minipage}[b]{0.248\textwidth}
            \centering
        \includegraphics[width=\textwidth]{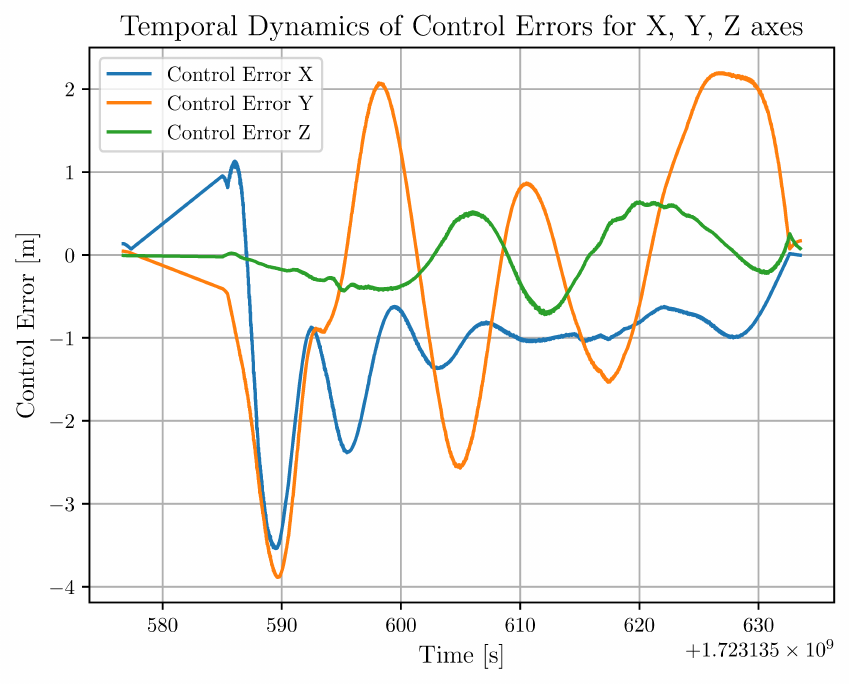}
        \end{minipage}
    \end{tabular}
    \caption{Comparative Analysis of UAV Path Following with NMPC in the Multiple Obstacle Scenario. The top-left image depicts in 3D, the desired path (blue dashed line) and the actual path (orange line), with the multiple obstacles illustrated by the green spheres
    (This applies to Figs.~\ref{fig:nmpc_path_tracking_obstacle} and~\ref{fig:combined_multi_obstacle}).}
\label{fig:combined_figure_multi_obstacle_4_simulator_results}
\end{figure}

To evaluate the NMPC controller performance in more realistic conditions, simulated HIL flight tests were conducted using the DJI Matrice 100 software. This approach involved connecting the UAV to a computer program that simulated the UAV dynamics based on the input commands and generates a full state, including GPS signals, enabling the UAV to react as though in flight despite remaining stationary on the desk. The controller was adaptable to different priorities. Table~\ref{tab:simulated_gps_flight_metrics} summarizes the simulator results. The control loop update times ranged from $\mathrm{0.0547}$ $\mathrm{s}$ to $\mathrm{0.0207}$ $\mathrm{s}$, indicating the system capability when it comes to real-time performance. The simulated HIL flight tests depicted in Figs.~\ref{fig:hexagon} and~\ref{fig:combined_figure_multi_obstacle_4_simulator_results}, compared with the initial simulations in Sec.~\ref{simulation_section} provided a more realistic UAV evaluation, revealing less precise path following because of real-world uncertainties like communication delays and sensor noise. Additionally, more iterations and adjustments were required for effective obstacle avoidance compared to simulations.

\vspace{-0.5cm}

\subsection{Real-World Performance: Outdoor Experimental Results}
\label{realworld_section}

\begin{figure}[htbp]
    \vspace{-0.46cm}
    \centering
    \captionsetup{font=footnotesize}
    \begin{tabular}{@{\hspace{-0.6mm}}c@{\hspace{-0.5mm}}c}
        \begin{minipage}[b]{0.248\textwidth}
            \centering
            \includegraphics[width=\textwidth]{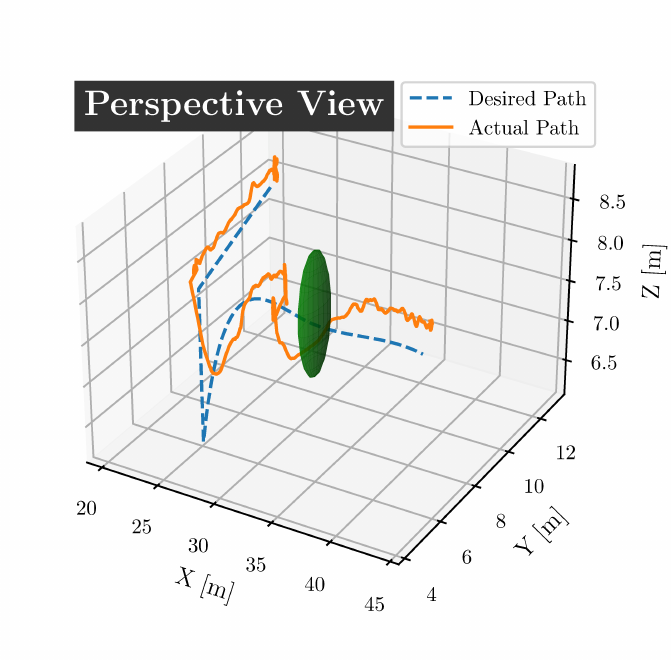}
        \end{minipage} &
        \begin{minipage}[b]{0.248\textwidth}
            \centering
            \includegraphics[width=\textwidth]{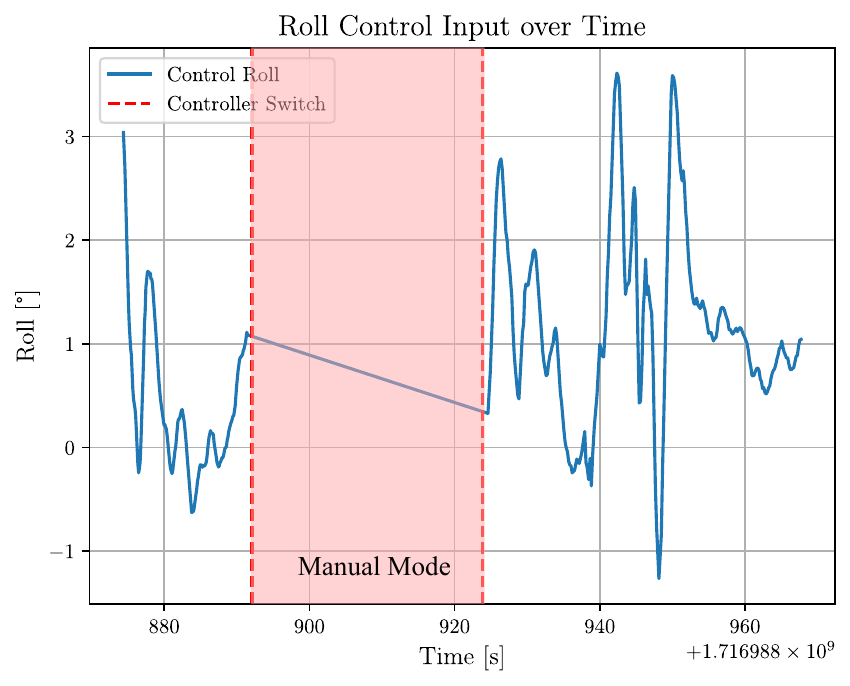}
        \end{minipage} \\
        \begin{minipage}[b]{0.248\textwidth}
            \centering
            \includegraphics[width=\textwidth]{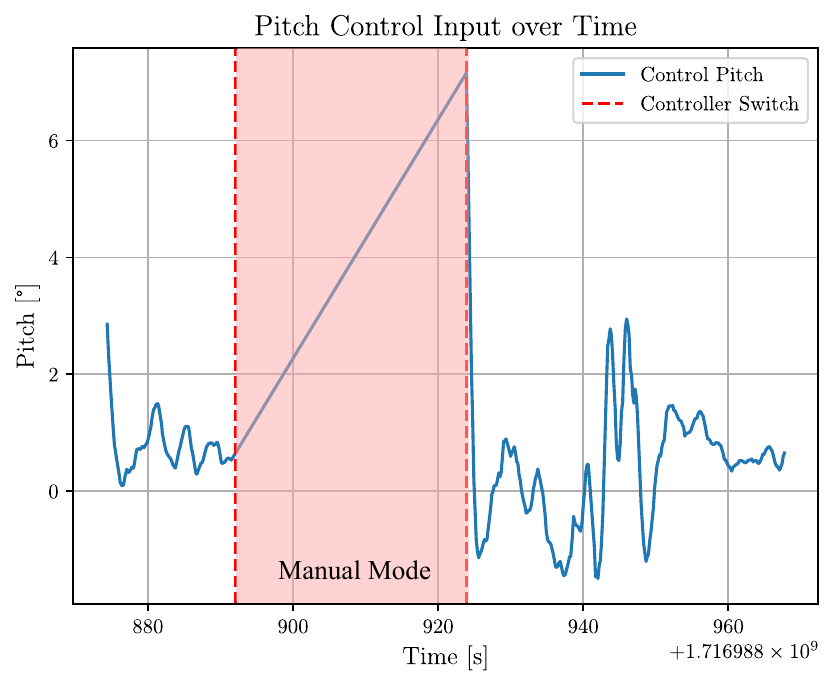}
        \end{minipage} &
        \begin{minipage}[b]{0.248\textwidth}
            \centering
            \includegraphics[width=\textwidth]{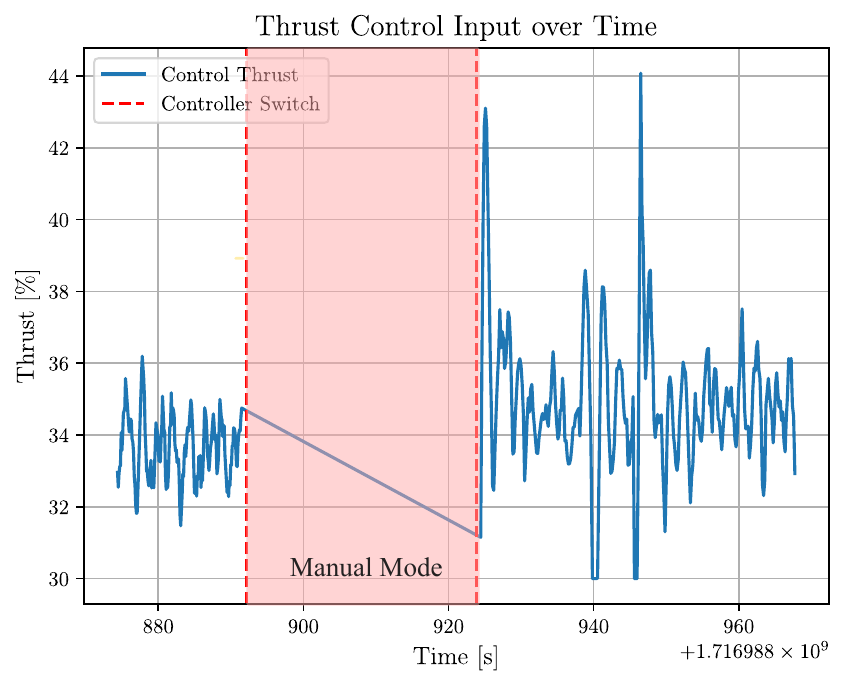}
        \end{minipage} \\
        \begin{minipage}[b]{0.248\textwidth}
            \centering
            \includegraphics[width=\textwidth]{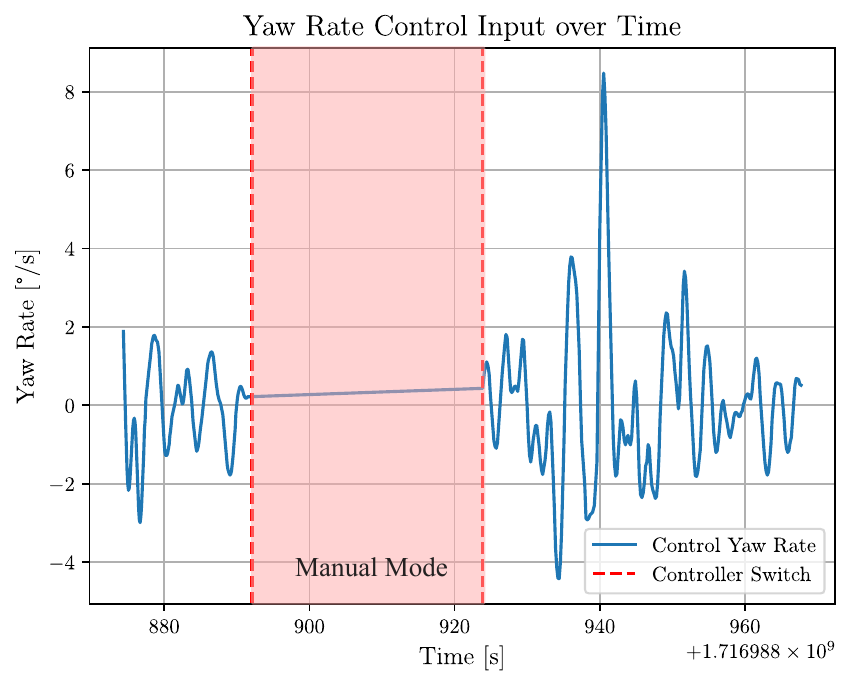}
        \end{minipage} &
        \begin{minipage}[b]{0.248\textwidth}
            \centering
            \includegraphics[width=\textwidth]{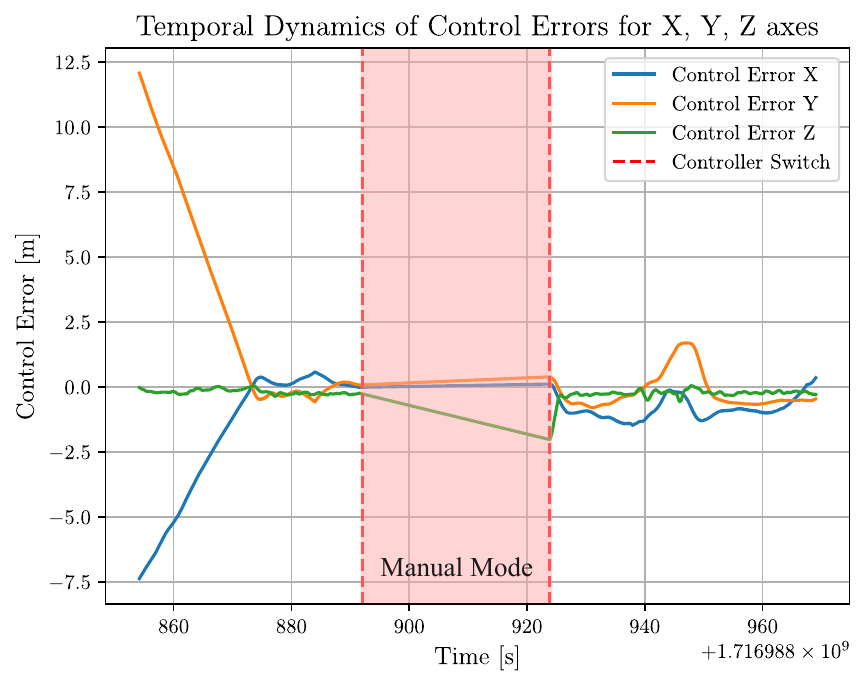}
        \end{minipage}
    \end{tabular}
    \caption{Comparative Analysis of UAV Path Following with NMPC in the Obstacle Scenario.}
    \label{fig:nmpc_path_tracking_obstacle}
    \vspace{-0.5cm}
\end{figure}

In Fig.~\ref{fig:nmpc_path_tracking_obstacle}, the UAV initially followed a B-spline path using NMPC. At $\mathrm{890}$ $\mathrm{s}$, it switched to manual control for precise positioning near obstacles, then back to NMPC with obstacle avoidance. 

\begin{table}[htbp]
\caption{Real-World Experimental Results}
\begin{center}
\footnotesize 
\begin{tabular}{|p{3.3cm}|p{1.4cm}|p{2.5cm}|}
\hline
\textbf{Metric}                          & \textbf{B-spline} & \textbf{One Obstacle} \\ \hline
\textbf{Navigation Time [s]}             & 74.48                 & 79.90                                \\ \hline
\textbf{Average Deviation [m]}           &  1.19                 & 1.43                                  \\ \hline
\textbf{Average Solver Iterations}               & 2.26                    & 2.72                                   \\ \hline
\textbf{Metric}                          & \textbf{B-spline} & \textbf{Multiple Obstacles} \\ \hline
\textbf{Navigation Time [s]}             & 72.93                 & 84.70                                \\ \hline
\textbf{Average Deviation [m]}           &  1.94                    & 2.48                                  \\ \hline
\textbf{Average Solver Iterations}               & 5.71                   & 7.22                                   \\ \hline
\end{tabular}
\label{tab:simulated_gps_flight_metrics_real_life}
\end{center}
\end{table}

\vspace{-0.5cm}

\noindent
This mode switch, marked by a spike seen in the red-shaded section in Fig.~\ref{fig:nmpc_path_tracking_obstacle}, occurred seamlessly in real time. The initial jump in altitude, observed during the manual switch and seen in Figs.~\ref{fig:nmpc_path_tracking_obstacle},~\ref{fig:combined_figure_multi_obstacle_3}, and ~\ref{fig:combined_multi_obstacle}, resulted from thrust adjustment based on real-time battery data analysis. The battery level was monitored, and thrust was calibrated accordingly to ensure stable altitude and to compensate for power fluctuations. The performance metrics in Table~\ref{tab:simulated_gps_flight_metrics_real_life} reflect the challenge of maintaining precise path following, with increased solver iterations and average deviation during obstacle-rich scenarios, highlighting the impact of real-world conditions.

\section{DISCUSSION}

If the gain factor \(\eta\) from Eq. (\ref{eq:eta_rolf}), which was chosen empirically, was not appropriately set, instability could occur in the control system, as discussed in the literature \cite{johansson2009stability}. This sensitivity to \(\eta\) remains a limitation, as its manual tuning can result in sub-optimal performance in various environments. During the HIL testing described in Sec.~\ref{HIL_section}, the controller was adaptable to different priorities. When velocity was heavily penalized, the UAV maintained precise positions with slow, steady movement,  resulting in accurate tracking of the hexagonal path. However, at higher speeds, where position control was less stringent, the path resembled a circle, Fig.~\ref{fig:hexagon}. Shifting the focus to velocity control allowed the UAV to avoid obstacles more quickly, with effective path guidance even at higher speeds. Despite variability in the control loop update times (as noted in Sec.~\ref{HIL_section}), the system maintained an update rate well above $\mathrm{0.1}$ $\mathrm{s}$, which was sufficient for stable UAV performance. Using control actions from the previously computed horizon as initial guess in the NMPC optimization ensured continuous trajectory tracking, even under higher computational loads, mitigating the potential impact of slower update rates. To improve computational efficiency, as described in Sec.~\ref{realworld_section}, obstacles were included in the cost function rather than as constraints, where they originally showed lower sampling rate performance. Additionally, faster sampling was achieved using a forward Euler method instead of the CasADi CVODES integrator or Runge-Kutta \cite{gustafsson1992control}.

\begin{figure}[htbp]
    \centering
    \captionsetup{font=footnotesize}
    \begin{tabular}{@{\hspace{-0.6mm}}c@{\hspace{-0.5mm}}c}
        \begin{minipage}[b]{0.248\textwidth}
            \centering
            \includegraphics[width=\textwidth]{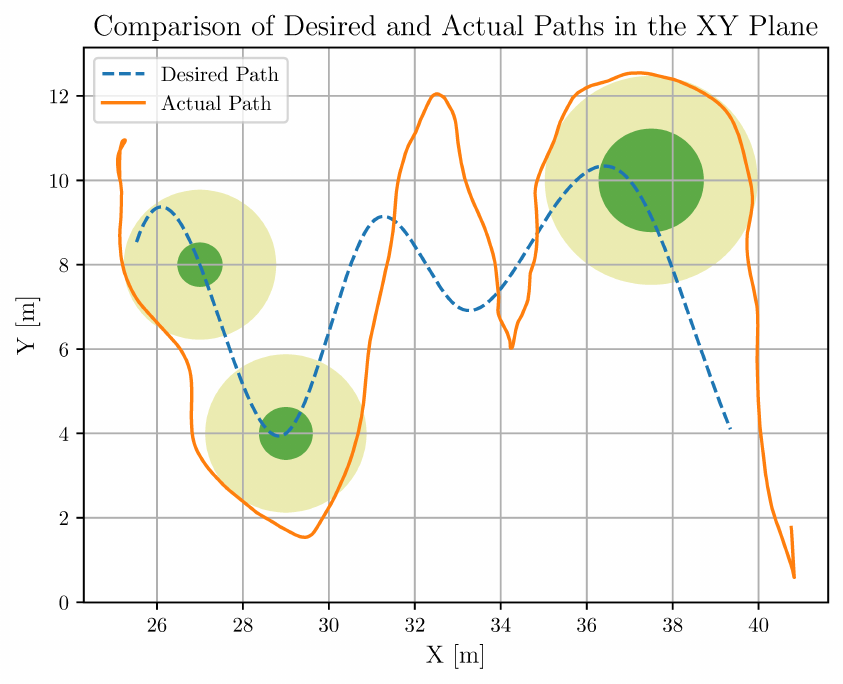}
        \end{minipage} &
        \begin{minipage}[b]{0.248\textwidth}
            \centering
            \includegraphics[width=\textwidth]{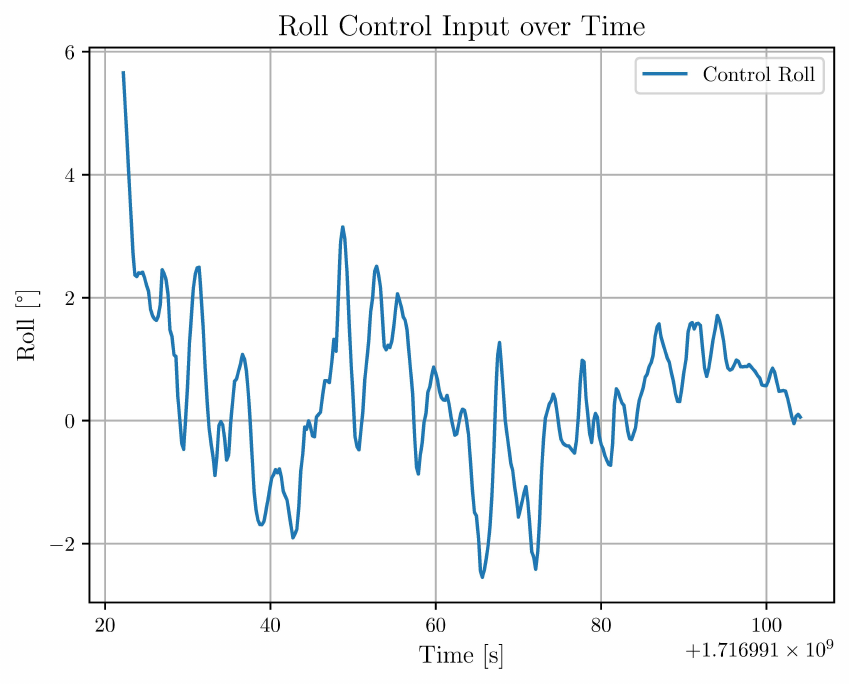}
        \end{minipage} \\
        \begin{minipage}[b]{0.248\textwidth}
            \centering
            \includegraphics[width=\textwidth]{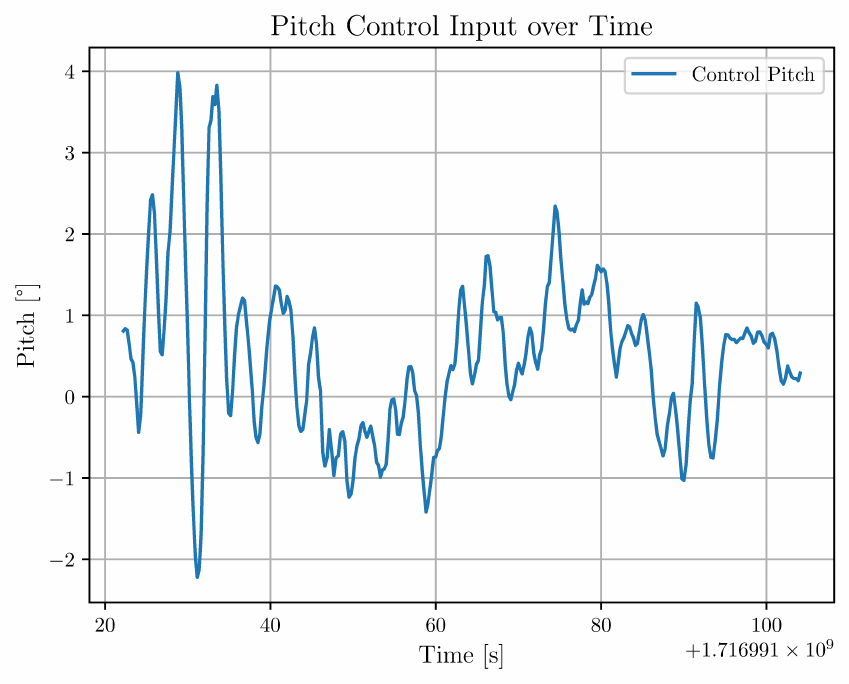}
        \end{minipage} &
        \begin{minipage}[b]{0.248\textwidth}
            \centering
            \includegraphics[width=\textwidth]{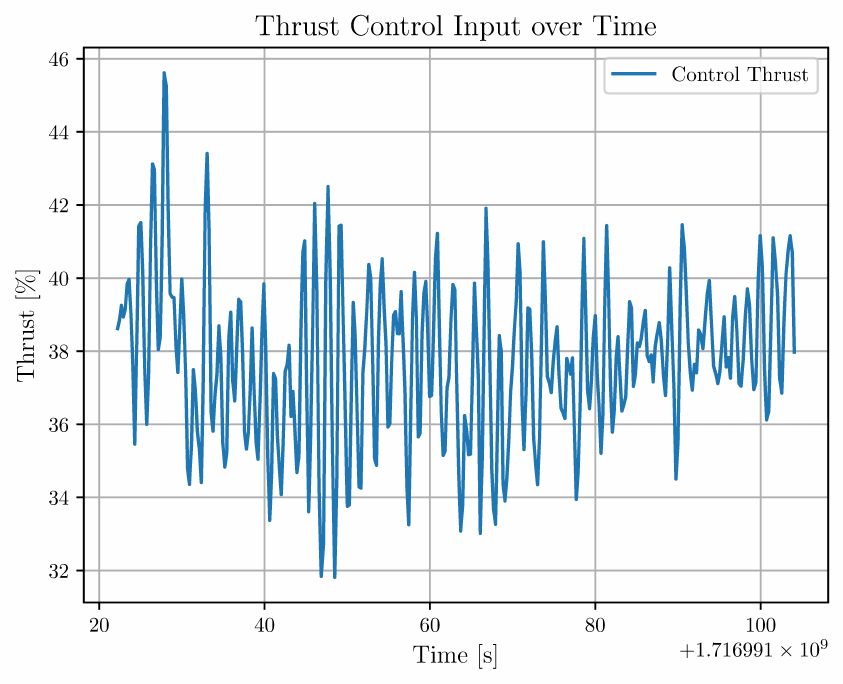}
        \end{minipage} \\
        \begin{minipage}[b]{0.248\textwidth}
            \centering
            \includegraphics[width=\textwidth]{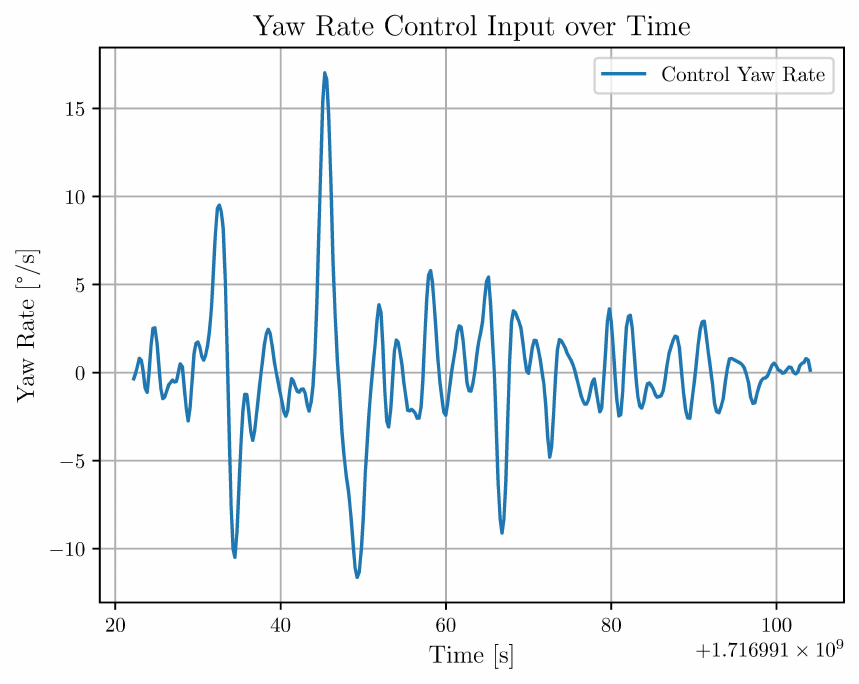}
        \end{minipage} &
        \begin{minipage}[b]{0.248\textwidth}
            \centering
            \includegraphics[width=\textwidth]{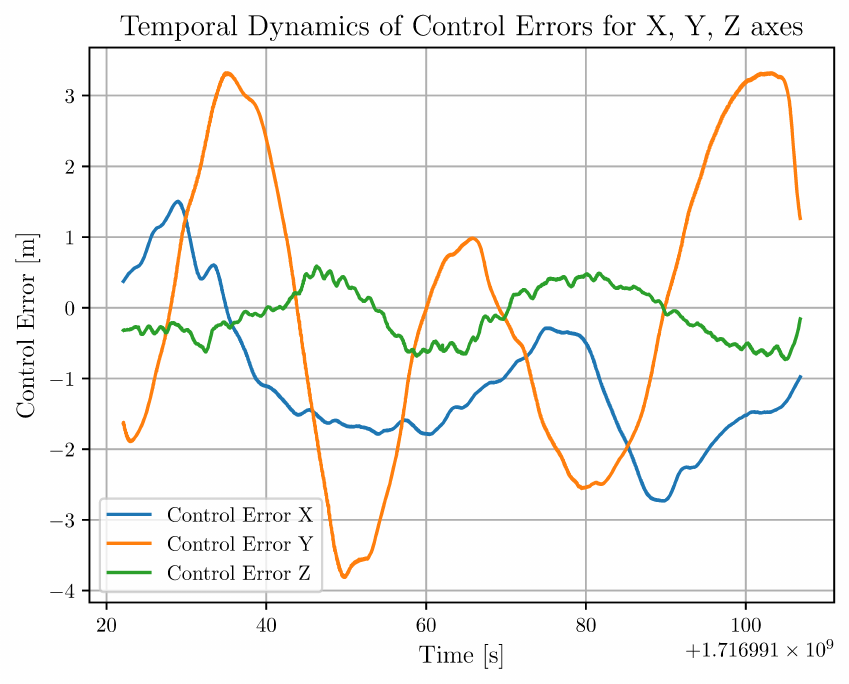}
        \end{minipage}
    \end{tabular}
    \caption{Comparative Analysis of UAV Path Following with NMPC in the Multiple Obstacle Scenario: The dashed blue line represents the desired B-spline path, while the orange line shows the UAV actual trajectory as it avoids obstacles (green circles with safety margins in pale yellow).}
    \label{fig:combined_figure_multi_obstacle_3}
    \vspace{-0.2cm}
\end{figure}
\begin{figure}[htbp]
    \centering
        \vspace{-0.4cm}
    \captionsetup{font=footnotesize}
    \begin{tabular}{@{\hspace{0.1mm}}c@{\hspace{-0.6mm}}c}
        \begin{minipage}[b]{0.225\textwidth}
            \centering
            \raisebox{2.0mm}{\includegraphics[width=\textwidth]{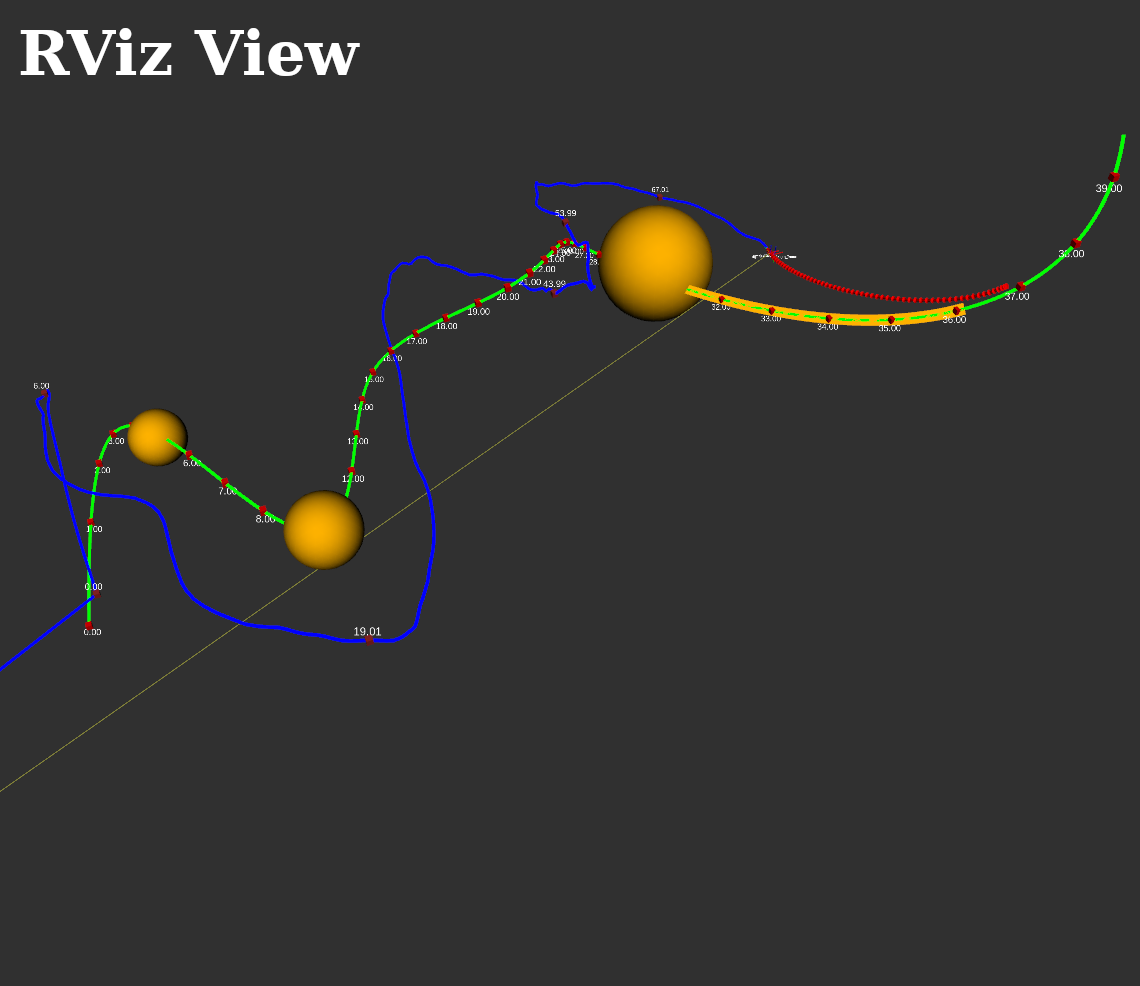}}
            \label{fig:rviz_multi_obstacle}
        \end{minipage} &
        \begin{minipage}[b]{0.268\textwidth}
            \centering
            \raisebox{-45mm}{\includegraphics[width=\textwidth]{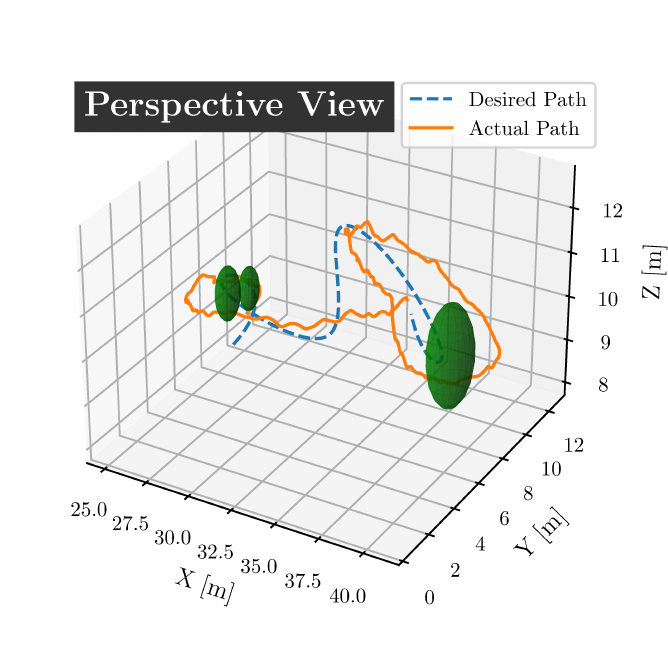}}
            \label{fig:nmpc_3d_path_tracking_multi_obstacle}
        \end{minipage}
    \end{tabular}
    \caption{Comparison of Desired and Actual Path Following in 3D with NMPC in the Multiple Obstacle Scenario: In the left image in RViz, the yellow sphere represents the obstacles, the green line is the desired trajectory, the blue line is the actual path taken by the UAV, the red line shows the prediction horizon, and the yellow line indicates the path being sent to the controller at the current time step.}
    \label{fig:combined_multi_obstacle}
    \vspace{-0.2cm}
\end{figure}

\vspace{-0.1cm}
\noindent
Safety distances were added, as shown in Figs.~\ref{fig:combined_figure_multi_obstacle_4_simulator_results} and~\ref{fig:combined_figure_multi_obstacle_3}, to minimize the risk of collisions. While the NMPC framework incorporates obstacle proximity directly into the cost function, ensuring effective collision avoidance, the safety distances serve as an additional precautionary measure to account for real-world uncertainties and sensor noise. Even if the UAV enters the safety margin, it has sufficient time to trigger corrective actions and avoid a collision. A longer prediction horizon increased the UAV ability to anticipate and navigate around obstacles, as seen in the improved path accuracy in Fig.~\ref{fig:combined_multi_obstacle}. Error analysis of the experimental results, shown in Figs.~\ref{fig:nmpc_path_tracking_obstacle} and~\ref{fig:combined_figure_multi_obstacle_3}, highlights that GPS inaccuracies caused minor deviations during obstacle avoidance. Despite these deviations, the UAV had effectively navigated the obstacles, validating the robustness of the NMPC approach under various conditions. 
A feature of our work is using the B-spline path representation, whereas previous works, such as \cite{lindqvist2020nonlinear}, focused on simpler indoor spline trajectories using a smaller UAV. In controlled indoor environments, our approach outperformed \cite{lindqvist2020nonlinear} by achieving smoother trajectory tracking with a deviation of $\mathrm{0.21}$ $\mathrm{m}$ compared to their reported deviations of $0.4 - 0.6$ $\mathrm{m}$. In outdoor environments, the larger deviations observed in our experiments are because of the increased complexity of testing in real-world conditions, which were not addressed by \cite{lindqvist2020nonlinear}, as their work was restricted to indoor experiments. Furthermore, \cite{kamel2017linear} performed all computations on the onboard NUC, while our framework uses offloading to an external workstation, while the onboard computer handled essential control tasks. In comparison to \cite{kamel2017linear}, we integrated the yaw rate into the NMPC, unlike their approach, which separated yaw rate control via a Proportional-Integral-Derivative (PID) controller. Additionally, our method offers a more generalized solution compared to the image-based obstacle avoidance of \cite{velasco2024visual}, as we directly incorporate UAV dynamics for obstacle avoidance, which improves robustness in varied environmental conditions.

\section{CONCLUSION AND FUTURE WORK}

The results validate the NMPC method’s ability to handle nonlinear dynamics, enforce constraints, and achieve accurate trajectory tracking across various scenarios, with the CasADi library supporting real-time optimization and ROS facilitating operational integration. While the system is well-suited for applications like precision agriculture, disaster response, and logistics, scaling it to environments with larger areas, more dynamic obstacles, and higher computational demands remains a challenge. The open-source code provided aims to support ongoing research and further development in UAV control. The code is designed to handle dynamic obstacles via ROS topics \cite{roswiki2024}, with real-time trajectory adjustments, and while not yet implemented, it allows for easy integration. Further refinements, including advanced prediction and obstacle handling algorithms in the control loop, would reduce latency and improve obstacle avoidance, particularly through better parameterization of the repulsive potential field constant.

\section{ACKNOWLEDGEMENT}
This work was made while the first author was financially supported by the project \emph{UAS@LU Autonomous Flight} conducted in cooperation between Lund University School of Aviation and Dept. Automatic Control, Lund University. The authors thank Björn Olofsson, Yiannis Karayiannidis, and Rolf Johansson for rewarding discussions on methodology and experiments.


\end{document}